# Visuo-tactile Diffusion Policy for Robotic Manipulation with Auxiliary Tactile Dynamics Prediction

Kaihong Ouyang
*School of Astronautics*
*Harbin Institute of Technology*
Harbin, Heilongjiang, China
25S104218@stu.hit.edu.cn

Zihao Su
*School of Astronautics*
*Harbin Institute of Technology*
Harbin, Heilongjiang, China
26S104180@stu.hit.edu.cn

Chengxuan Zheng
*School of Astronautics*
*Harbin Institute of Technology*
Harbin, Heilongjiang, China
26B904082@stu.hit.edu.cn

Weiyi Lai
*School of Astronautics*
*Harbin Institute of Technology*
Harbin, Heilongjiang, China
26S104138@stu.hit.edu.cn

Dong Zhou*
*School of Astronautics*
*Harbin Institute of Technology*
Harbin, Heilongjiang, China
dongzhou@hit.edu.cn
**Corresponding author*

**Abstract—Long-horizon robotic assembly requires reliable perception and action across grasping, alignment, and insertion. Vision captures global object pose and scene context, whereas fingertip touch provides local contact cues that are especially useful as contact states change. We present a visuotactile diffusion policy that combines four-step histories from two camera views and two fingertip tactile images to condition action-sequence generation. A recurrent temporal representation integrates recent observations from both modalities. To encourage it to capture how contact evolves under action, an auxiliary objective predicts fingertip-image changes two steps ahead from the temporal representation and demonstrated actions. This objective is used only during training and adds no inference-time computation. Across three independent training seeds in MuJoCo assembly experiments, the full model achieves an average success rate of 59.0% ± 2.6%, improving on the visual-only baseline (42.7% ± 1.5%) by 16.3 percentage points.**

## I. INTRODUCTION

The widespread application of embodied intelligent robots in industrial manufacturing has made autonomous perception and action in complex physical environments an important research focus. These capabilities are also important for planetary exploration, where autonomous robots must plan and execute under limited supervision [1], [2], [3], [4]. Assembly requires a robot to complete interdependent steps such as grasping, alignment, and insertion, placing high demands on environmental perception, contact-state estimation, and action selection. In long-horizon tasks, pose deviations or grasping errors introduced early in an episode may propagate to later stages, making reliable execution of the complete task difficult. Accordingly, using multi-modal observations to infer manipulation state and generate coherent action sequences is central to improving robotic assembly capability.

Vision and touch provide complementary information for robotic manipulation. Vision captures the object's position, orientation, and surrounding workspace, whereas tactile sensing reveals local contact changes relevant to grasp assessment and fine adjustment. However, relying on either modality alone has limitations throughout a continuous assembly process. As the manipulation state evolves, a single observation stream may not reliably distinguish newly established contact, unexpected object slip, or small pose corrections, especially when the object or gripper occludes the camera view. Jointly using the recent history of visual and tactile observations links global object motion with local contact-state changes, providing temporal cues for more reliable action selection during grasping, alignment, and insertion.

These time-varying multimodal cues must then be translated into coordinated action sequences that can be revised as contact conditions change. Diffusion policies provide a framework for modeling such sequences in imitation-based robotic manipulation. Diffusion Policy [7] formulates a visuomotor policy as a conditional denoising diffusion process, generates action sequences through iterative denoising, and executes them in a receding-horizon manner. Building on this framework, 3D-ViTac [8] fuses visual and tactile information in a unified three-dimensional representation while preserving spatial structure and correspondence for diffusion-policy learning. Reactive Diffusion Policy [9] instead combines slow latent-space action planning with fast tactile-feedback control to improve responsiveness in contact-rich manipulation. These studies show that multimodal representations and feedback mechanisms are important design choices for visuotactile policies and motivate further investigation of how visuotactile observations can support continuous action generation.

A visuo-tactile policy must learn the relationship between local observations and effective actions from training data. Changes in tactile images depend both on the current contact state and on robot actions. Learning the tactile consequences of actions can therefore provide additional supervision for the observation representation. Large-scale collections of robot experience further demonstrate the value of diverse training data [10]. Moreover, execution errors may place the robot in states that are poorly covered by the original demonstrations, whereas teacher corrections can supply action labels for such states. Intervention-based imitation learning collects and emphasizes corrective demonstrations to help policies pass manipulation bottlenecks [15].

To address these issues, we develop a visuo-tactile diffusion policy with auxiliary tactile-dynamics prediction (Fig. 1). Two cameras provide visual features, a convolutional network encodes rendered dual-fingertip tactile-proxy images, and a GRU aggregates four frames of visuotactile features into a temporal representation for action generation. This representation is concatenated with the visual history and supplied to a conditional one-dimensional U-Net that generates action sequences through iterative denoising. During training, an auxiliary head predicts tactile changes two steps into the future from the temporal representation and demonstration actions. The training set combines expert demonstrations, teacher corrections, and robot experience. Ablations over visual-only, visuo-tactile, and auxiliary prediction variants assess the contribution of each component to complete-task execution.

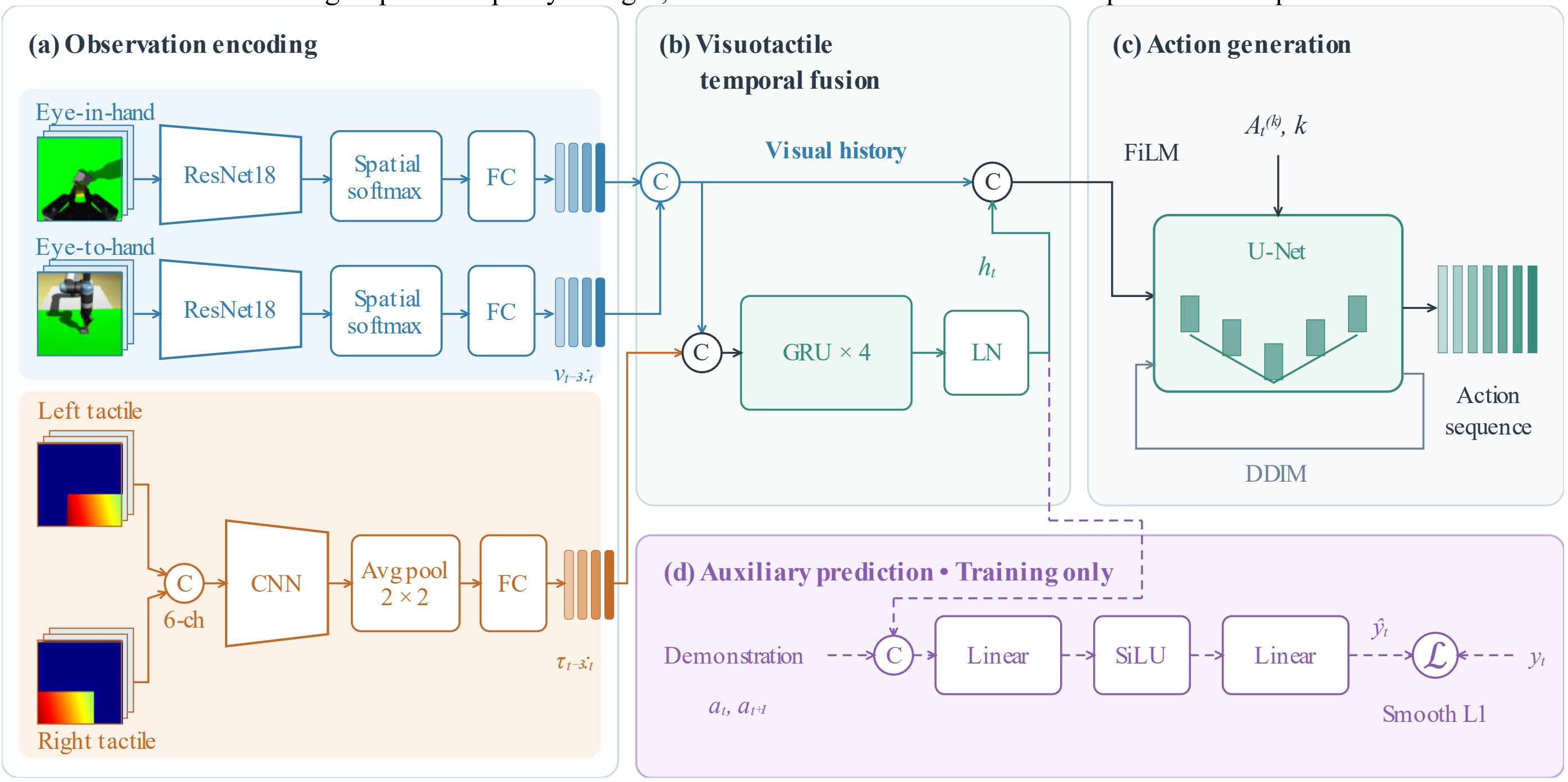

Fig. 1. Visuotactile diffusion policy with auxiliary tactile-dynamics prediction. (a) Encoding the dual-view visual history and dual-fingertip tactile history; (b) GRU-based aggregation of frame-wise visuotactile features, whose output is concatenated with the visual history; (c) conditional diffusion model for action-sequence generation; and (d) training-only prediction of tactile change two steps into the future using demonstration actions.

## II. Related Work

### A. Long-Horizon Assembly and Phase-Aware Policies

Long-horizon assembly tasks typically comprise interdependent stages such as grasping, alignment, and insertion. As a task progresses, the geometric constraints, contact states, and perceptual requirements change accordingly. Phase information provides task context for action selection and can determine how different sensing modalities are used. FoAR [16] predicts future contact probabilities to dynamically regulate the fusion of visual and force/torque information, enabling different perceptual conditions in contact and noncontact states. PhaForce [17] further introduces an explicit distribution over task phases, uses contact and phase information to modulate visual–force fusion in diffusion planning, and guides high-frequency residual corrections.

This work focuses on temporal visuotactile representation and action generation throughout a continuous assembly process. The model jointly encodes visual and fingertip observation histories, aggregates recent interaction information with a temporal network, and conditions a diffusion action model on the resulting representation. Future tactile changes provide auxiliary supervision so that the temporal representation supports both action prediction and prediction of local interaction changes. The policy executes in a receding-horizon manner, continually generating subsequent actions from updated observations, and is evaluated using full-task success rate.

### B. Tactile Perception and Visuo-tactile Fusion

Optical tactile sensors record image changes caused by local contact, enabling tactile features to be extracted using visual encoders. DIGIT [5] introduced a compact, high-resolution tactile sensor for in-hand manipulation, whereas TACTO [6] provides tactile simulation and rendering tools for such sensors. These works support image-based local contact perception and its application to robotic manipulation. We use image-based fingertip proxy observations in simulation to study the fusion of local interaction information with scene vision; their generation and representation are defined in Section III.

Visuotactile fusion methods exploit the complementarity of the two modalities through spatial representations, cross-modal associations, and adaptive weighting. 3D-ViTac [8] maps vision and touch into a unified three-dimensional representation, preserves their spatial structure and relationships, and combines the representation with a diffusion policy for fine-grained manipulation. AdapTac [18] uses the current observation and predicted future forces to guide attention and adaptively weight visual and tactile features. In this work, temporal aggregation of visual and tactile features constructs the action condition, while an auxiliary task that predicts future tactile-image changes supervises the learning of local interaction representations.

### C. *Diffusion Policies and Teacher Corrections*

Denoising diffusion models represent data distributions by learning a reverse denoising process [11]. Diffusion Policy [7] applies this idea to robot action-sequence generation and uses receding-horizon execution to continually revise actions from updated observations; DDIM [12] provides a basis for reducing the number of denoising steps. Direct policy optimization has also been investigated for diffusion policies [13], while open-source frameworks such as LeRobot support reproducible learning-based robotics workflows [14]. This work uses a conditional one-dimensional U-Net to generate action sequences and combines the action-denoising loss with a tactile-dynamics prediction loss during supervised training.

Intervention Weighted Regression (IWR) [15] increases the weight of corrective demonstrations, strengthening the learning of corrective behavior while retaining non-intervention data. Following this idea, teacher-correction data are used together with the original expert demonstrations and robot experience for supervised diffusion-policy training, and teacher-correction samples are drawn with higher probability. The teacher-correction data include action labels generated from states recorded immediately before takeover, providing corrective demonstrations for the corresponding states.

## III. METHOD

### A. *Task and Observation Representation*

The simulated task requires the robot to approach a target, establish a grasp, transport the object to a specified target region, and align it for assembly. Actions are represented by seven dimensions: three translational increments, three rotational increments, and one gripper-control value. Each action prediction uses the four most recent observations, ordered chronologically. Visual observations are provided by a scene camera and a hand camera, while local fingertip observations comprise rendered RGB tactile-proxy images from the left and right fingertips. Fig. 2 shows the two camera views and dual-fingertip tactile-proxy observations at the same time step.

The stored camera history has dimensions 256 × 256 × 24, and the stored fingertip history has dimensions 24 × 64 × 64. The 24 channels correspond to four time steps, two images, and three color channels. Camera images are scaled to [0, 1], resized to 128 × 128 using antialiased bilinear interpolation, and standardized using channel means and standard deviations estimated in advance from the training samples. Fingertip images retain a resolution of 64 × 64; the left and right tactile images at each time step are concatenated along the channel dimension to form a six-channel input. Training and inference use the same fingertip-image quantization and normalization procedure. Channel means and standard deviations are estimated from the training samples, and the standard deviation is lower-bounded at 0.02.

Expert demonstrations and mixed-source data are organized and preprocessed in the same manner to provide a uniform policy input. Fingertip observations are RGB images representing local geometry and interaction information; the images in the historical expert demonstrations are generated by rendering local geometry. We use these tactile-proxy observations together with scene vision for action prediction. Throughout this paper, tactile input and tactile-dynamics prediction refer to these rendered fingertip images and their temporal changes. They are not measurements from a physical tactile sensor, and their pixel values do not quantify calibrated contact forces. Simulated tactile sensors differ from physical sensors, and the rendered images are generally cleaner and contain less noise than images acquired from real sensors.

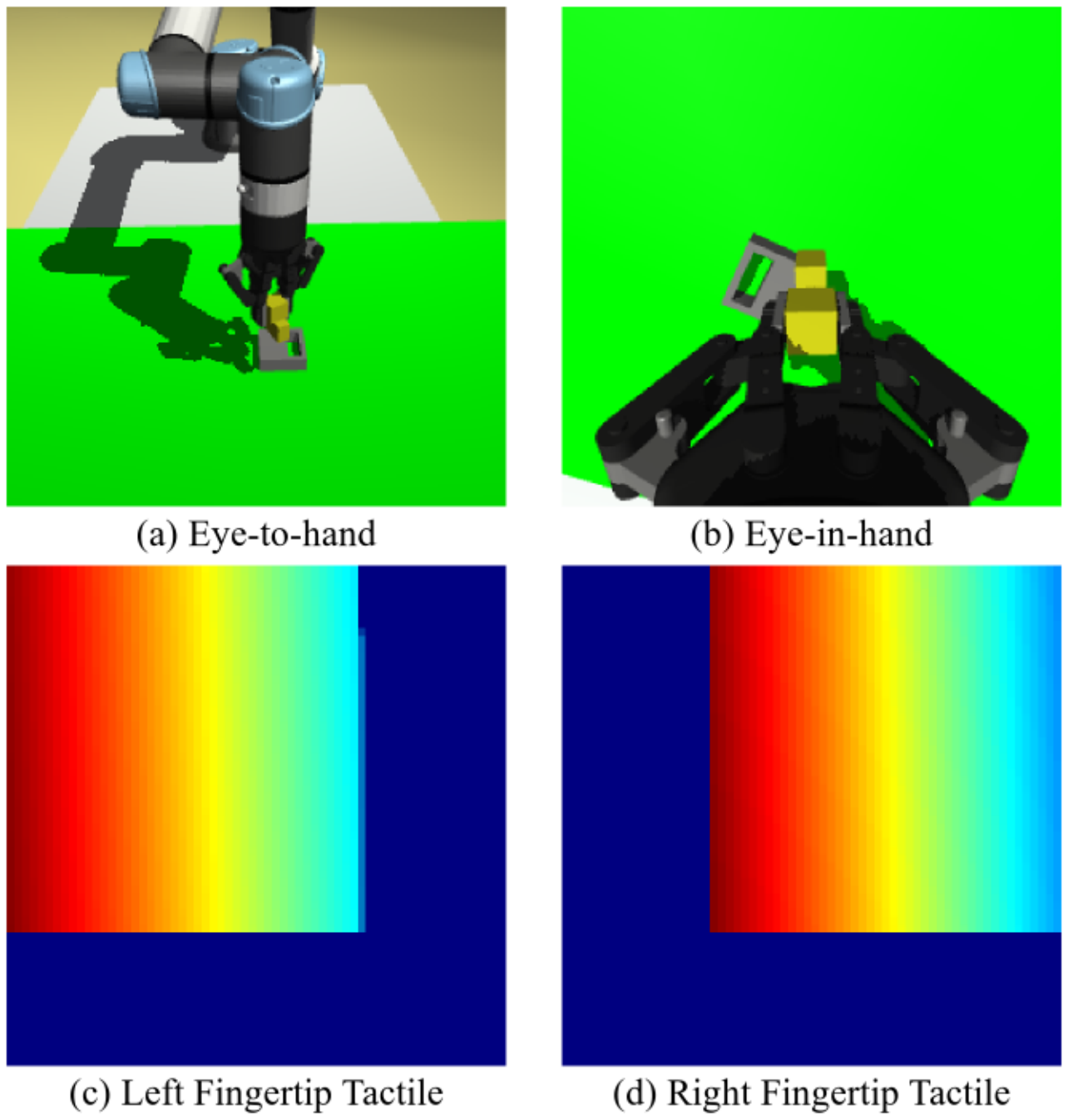

Fig. 2. Visual and fingertip observations at the same time step. The top row shows the Eye-to-hand and Eye-in-hand camera images, and the bottom row shows the left and right fingertip tactile-proxy images. Colors in the fingertip images encode local geometry and interaction information and do not correspond to calibrated contact forces.

### B. *Observation Encoding and Visuotactile Fusion*

Fig. 1 shows the overall architecture. The visual branch consists of two independently parameterized ResNet-18 encoders [19] for the scene and hand images. Each encoder applies spatial softmax [20] to extract the expected two-dimensional coordinates of 32 keypoints, followed by a linear projection and activation that produce a 64-dimensional feature for each image. The encoder for a given view shares parameters across the four observation times. The visual backbones are initialized with ImageNet-pretrained weights and first trained by behavior cloning as part of the visual-only diffusion policy; their parameters and normalization statistics are then frozen during subsequent visuotactile training.

At time step i, the two visual features are concatenated into the 128-dimensional vector $v_i$. To retain the input interface of the visual base policy, a state term fixed at zero is appended at each time step, yielding $z_i^v = [0;\ v_i\ ]$. Concatenating the representations from the four time steps produces the 516-dimensional visual-history condition:

$$c_t^v = \mathrm{concat}(z_{t-3}^v,\ z_{t-2}^v,\ z_{t-1}^v,\ z_t^v). \quad (1)$$

At each time step, the tactile branch receives the six-channel fingertip image and applies three convolutional layers with stride 2. The layers have 32, 64, and 128 channels and kernel sizes of 5, 3, and 3, respectively. Each layer uses group normalization with eight groups and a SiLU activation. A 2 × 2 adaptive average pooling layer followed by a linear projection produces the 128-dimensional tactile feature $\tau_i$. After layer normalization, the visual feature is concatenated with the tactile feature to form the 256-dimensional input $u_i = [\mathrm{LN}(v_i);\ \tau_i]$.

The temporal network uses a GRU with hidden dimension 128 to aggregate the joint features from the four most recent time steps. The output at the final time step is layer-normalized to obtain the temporal representation $h_t$:

$$h_t = \mathrm{LN}(\mathrm{GRU}\ (u_{t-3:t})_{\mathrm{last}}). \quad (2)$$

The GRU is computed independently within each four-frame observation window, with its hidden state initialized at the start of the window. The final 128-dimensional temporal representation is concatenated with the 516-dimensional visual-history condition to form the 644-dimensional condition for action generation:

$$c_t = \mathrm{concat}(c_t^v,\ h_t) \in \mathbb{R}^{644}. \quad (3)$$

Three distinct concatenation operations are used: the left and right fingertip images are concatenated along the pixel-channel dimension; the per-frame visual and tactile features are concatenated along the feature dimension and supplied to the GRU; and the four-frame temporal representation is concatenated with the visual-history condition before entering the action-denoising network. The visual history is flattened in a fixed order, whereas the visuotactile temporal representation is learned by the GRU. Together, they provide the observation condition required for action prediction.

The action generator is a conditional one-dimensional U-Net that performs convolution along the temporal dimension of the action sequence. Diffusion step k is encoded as a 128-dimensional vector and concatenated with the 644-dimensional observation condition, producing a 772-dimensional conditioning input for each conditional residual block. Each block uses feature-wise linear modulation (FiLM) [21] to generate per-channel scale and bias terms that modulate the intermediate action features:

$$\tilde{h}_\ell = \gamma_\ell(c_t, k) \odot h_\ell + \beta_\ell(c_t, k). \quad (4)$$

Here, $h_l$ denotes the action feature in residual block ℓ. The scale and bias are broadcast along the temporal dimension of the action sequence. The U-Net channel widths are 256, 512, and 1,024; the residual blocks use a kernel size of 5 and group normalization with eight groups. The model is initialized from the trained visual-only policy. The original conditioning-map parameters are retained, and the newly introduced conditioning weights corresponding to the temporal feature are initialized to zero.

### C. Auxiliary Tactile-Dynamics Prediction

The auxiliary task predicts action-conditioned changes in the rendered fingertip proxy images. It does not estimate calibrated physical contact dynamics. Let $T_t$ denote the current six-channel fingertip image and $T_{t+2}$ the image recorded after two actions have been executed. Their difference is computed, adaptively average-pooled to 4 × 4 per channel, and flattened to form the 96-dimensional tactile-change target:

$$d_t = \mathrm{vec}(\mathrm{Pool}_{4\times 4}(T_{t+2} - T_t)) \in \mathbb{R}^{96}. \quad (5)$$

The target is standardized using the per-dimension mean $\mu_d$ and standard deviation $\sigma_d$ estimated from the training samples, yielding $y_t = (d_t - \mu_d)/\sigma_d$. The standard deviation is lower-bounded at 0.001. This target describes fingertip-image changes over the next two action steps and can be constructed automatically from existing trajectories.

The auxiliary prediction head receives the temporal representation $h_t$ and the normalized first two demonstration actions $a_t$ and $a_{t+1}$. Its input dimension is 142. The prediction head is a two-layer multilayer perceptron with a SiLU activation between the layers and outputs an estimate of the standardized tactile change:

$$\hat{y}_t = g(\mathrm{concat}(h_t,\ a_t,\ a_{t+1})). \quad (6)$$

During training, a Smooth L1 loss constrains the prediction to the target, so the temporal representation also receives tactile-dynamics supervision. Demonstration actions are supplied only to the auxiliary head; the action-generation condition $c_t$ is computed entirely from current and historical observations. At deployment, only the observation encoders, GRU, and action-denoising network are used; the auxiliary head does not participate in action inference.

### D. Action Diffusion and Joint Training Objective

Each training window contains eight actual actions aligned with the latest observation. The action network uses a length-16 sequence tensor $A_t$. The eight valid actions occupy zero-based indices 3–10, while the first three and final five positions are padded with boundary actions. During training, an index k is sampled from 100 diffusion-noise levels and Gaussian noise is added according to a cosine schedule:

$$A_t^{(k)} = \sqrt{\bar{\alpha}_k} A_t + \sqrt{1-\bar{\alpha}_k}\ \varepsilon. \quad (7)$$

The conditional one-dimensional U-Net takes $A_t^{(k)}$, diffusion step k, and observation condition $c_t$ to predict the added noise. Let $\hat{\varepsilon}_j$ denote the predicted noise at position j, and let mask $m_j$ equal 1 at indices 3–10 and 0 elsewhere. The action-denoising loss is then:

$$L_{\text{diff}} = \mathrm{E}[(1/56)\, \Sigma_{j=0}^{15} m_j \, \|\varepsilon_j - \hat{\varepsilon}_j\|_2^2]. \quad (8)$$

The loss is averaged over the seven components of the eight valid actions, for a total of 56 scalar targets. The complete model is trained using the sum of the action-denoising and auxiliary-prediction losses:

$$L = L_{\text{diff}} + \lambda_{\text{aux}}\, \mathrm{SmoothL1}\, (\hat{y}_t, y_t). \quad (9)$$

The auxiliary-loss weight $\lambda_{aux}$ is set to 0.05. Training also records the loss of a constant predictor that assumes no future tactile change, allowing the auxiliary predictor to be compared with a simple zero-change estimate.

### *E. Teacher Corrections and Training Procedure*

The training data comprise three sources: the original expert demonstrations, corrective segments executed during teacher intervention, and segments autonomously executed by the robot policy after the teacher relinquishes control. The teacher is a scripted policy used only during data collection and has access to the full simulator state, whereas the robot policy receives only the visual and tactile observations defined above. Data are collected using manipulated objects with multiple geometries, recording visual observations, fingertip observations, and corresponding actions for each shape. Trajectories from all object shapes are pooled into a single dataset and jointly used for policy training.

During a mixed-control episode, the robot first performs the task autonomously while the scripted teacher continuously monitors execution using the full environment state. When a predefined deviation criterion is satisfied, indicating that the autonomous behavior has clearly departed from the task objective, the teacher temporarily takes control and executes a fixed number of corrective actions to guide the robot back toward a recoverable state. Control is then returned to the robot, and the teacher may intervene again if another deviation occurs within the same episode. The simulator state available to the teacher is used only for intervention triggering and corrective-action generation and is never provided to the robot policy.

A mixed-control episode is included in the dataset only if it ultimately succeeds and contains complete temporal windows satisfying the training requirements. The dataset retains the actions actually executed in the environment during both autonomous execution and teacher takeover. Consequently, the policy is trained not only on nominal expert behavior, but also on states encountered before and after autonomous deviations together with the corresponding recovery behavior.

From the available temporal windows, 358,499 samples are drawn for training and 41,601 samples for validation. The target sampling proportions of teacher-correction, expert-demonstration, and robot-experience data are 50.0%, 32.7%, and 17.3%, respectively. To account for stochasticity in policy optimization, the complete training procedure is repeated using three independent random seeds. For each seed, a visual-only diffusion policy is first trained as the corresponding base policy. The Vision+Touch, Vision+Aux, and Vision+Touch+Aux variants are then initialized from the seed-matched visual-only policy. Within each random seed, all model variants use the same training and validation data, sample order, and corresponding random settings to ensure a controlled comparison, while the network architecture and all training hyperparameters are kept fixed across seeds and model variants.

After the visual encoders are trained as part of the visual-only base policy, their parameters and normalization statistics are frozen. Visual features are subsequently precomputed and cached, while the tactile encoder, temporal network, and action-denoising network are optimized according to the corresponding model configuration. AdamW is used with zero weight decay, a gradient-norm clipping threshold of 1.0, and an exponential-moving-average decay of 0.98. For each model and random seed, the EMA checkpoint is selected according to the action-denoising loss on the fixed validation set.

At inference time, a length-16 action tensor is initialized from Gaussian noise and denoised using 10 DDIM steps. The eight valid actions at indices 3–10 are extracted and transformed back to the original action scale. After these eight actions are executed, the robot acquires an updated four-frame observation history and generates a new action sequence, resulting in receding-horizon execution.

## IV. Experiments

### *A. Experimental Setup*

The simulation environment is built in MuJoCo and uses a six-degree-of-freedom UR5 manipulator. MuJoCo simulates the dynamics and contact interactions among the robot arm, end-effector gripper, manipulated object, and environment. The experiments include manipulated objects with multiple geometries, all of which are covered during data collection and policy training. The same task procedure, observation format, and 7-dimensional actions are used across object shapes.

To account for both training stochasticity and evaluation randomness, each model variant is trained using three random seeds. For each training seed, every corresponding policy is evaluated over 100 episodes. Episode-wise paired evaluation is used within each seed: corresponding episodes for different methods use identical environment-reset seeds and policy-sampling seeds, and the generated initial states are verified to match after reset. The same set of 100 evaluation episodes is used for all model variants within a training seed. For each seed, success rate is the number of successful episodes divided by 100, expressed as a percentage; thus, each additional success changes the rate by one percentage point. Unless otherwise stated, all methods use the same environment-reset protocol and a 50-step simulation-settling period.

Each policy takes the four most recent observations as input, replans after executing the first eight predicted actions, and uses 10 DDIM denoising steps during inference. Success rate is computed independently for each training seed. Table II reports the mean and sample standard deviation across the three training seeds; the ± values are standard deviations, not confidence intervals.

For uncertainty analysis, the training seed is the unit of replication. We pair the full model and visual-only baseline by training seed and compute the mean success-rate difference. An exploratory 95% Student-t confidence interval is calculated as the mean paired difference ± 4.303 times its sample standard deviation divided by √3 (two degrees of freedom). This interval assumes independent, approximately normally distributed seed-level differences; episode count is not substituted for the number of independent training runs.

To separately examine the effects of online tactile feedback and auxiliary tactile-change supervision, we construct the following comparisons. The full model, denoted Vision+Touch+Aux, uses visual observations, tactile observations, and the tactile-change prediction loss. The Vision+Aux variant retains the auxiliary task but fixes the tactile encoder input to zero; consequently, this branch provides the GRU with a sample-invariant feature independent of the current tactile measurement. Tactile data therefore serve only as supervision for the auxiliary loss during training and are not used as policy observations during execution, so action generation depends only on vision. The Vision+Touch variant uses both visual and tactile observations but removes the tactile-change prediction loss, whereas the Vision-only variant uses visual observations alone.

TABLE I. ABLATION SETTINGS

| Model | Visual Input | Tactile Input | Auxiliary Tactile-Change Supervision |
|---|---|---|---|
| Vision+Touch+Aux | ✓ | ✓ | ✓ |
| Vision+Touch | ✓ | ✓ | – |
| Vision+Aux | ✓ | – (zero input) | ✓ |
| Vision-only | ✓ | – | – |

## B. *Performance Comparison*

TABLE II. ABLATION RESULTS ACROSS THREE TRAINING SEEDS UNDER MATCHED EVALUATION CONDITIONS

| Model | Seed 1 | Seed 2 | Seed 3 | Success Rate |
|---|---|---|---|---|
| Vision+Touch+Aux | **62%** | **58%** | **57%** | **59.0 ± 2.6%** |
| Vision+Touch | 33% | 34% | 34% | 33.7 ± 0.6% |
| Vision+Aux | 41% | 42% | 44% | 42.3 ± 1.5% |
| Vision-only | 44% | 43% | 41% | 42.7 ± 1.5% |

As shown in Table II, the full Vision+Touch+Aux model achieves an average success rate of 59.0% ± 2.6%, compared with 42.7% ± 1.5% for Vision-only, 33.7% ± 0.6% for Vision+Touch, and 42.3% ± 1.5% for Vision+Aux. The full model improves the average success rate over the visual-only baseline by 16.3 percentage points and exceeds all other variants in each of the three training seeds. Matched initial conditions and policy-sampling settings provide a common evaluation basis for comparing the performance of the model variants.

The full-model gains over Vision-only are 18, 15, and 16 percentage points for Seeds 1–3, respectively. Their mean is 16.33 percentage points, with a sample standard deviation of 1.53 and an exploratory 95% Student-t confidence interval of [12.5, 20.1] percentage points. An exact two-sided sign test for the three positive differences gives $p = 0.25$. The gain ranges from 15 to 18 percentage points, showing that the full model improves task success over the visual-only baseline in every training seed. These paired comparisons complement the mean success rates in Table II and support the benefit of combining tactile observations with auxiliary prediction for the simulated assembly task.

Table II distinguishes the effect of directly adding fingertip observations from that of adding auxiliary tactile-change supervision. Vision+Touch achieves 33.7% ± 0.6%, below the 42.7% ± 1.5% obtained by Vision-only, with seed-wise decreases of 11, 9, and 7 percentage points. Thus, directly supplying tactile input does not produce a performance gain in this experiment. With the same visual and fingertip observation histories, adding the auxiliary objective increases success to 59.0% ± 2.6%, improving on Vision+Touch by 25.3 percentage points and on Vision-only by 16.3 percentage points. The gains over Vision+Touch are 29, 24, and 23 percentage points across the three seeds. This comparison motivates examining how auxiliary supervision helps the policy use the available local information.

Adding tactile observations gives the policy access to local interaction information, but using this information effectively requires a representation that relates local states to robot actions. In Vision+Touch, the tactile encoder and GRU are supervised solely by the action-denoising loss. The relationship between current fingertip observations, executed actions, and subsequent local changes is therefore learned implicitly. The lower success rate of this variant suggests that, in the present experiments, supplying tactile observations alone is insufficient to improve task execution.

The auxiliary task provides explicit supervision for this relationship by predicting fingertip-image changes two action steps ahead, conditioned on the shared temporal representation and the first two demonstrated actions. Because similar fingertip appearances can evolve differently depending on the action and recent interaction history, this prediction encourages the representation to retain information relevant to the subsequent changes. The prediction loss directly supervises the tactile encoder and GRU, helping link local observations to their evolution under action. Both variants use the same four-frame histories and GRU-based fusion, so the added supervision acts on the existing temporal representation. It guides learning toward local interaction changes while complementing the global motion and task context captured by vision.

The Vision+Aux comparison highlights the contribution of current tactile observations when learning to predict local changes. Its success rate is 42.3% ± 1.5%, close to Vision-only, with seed-wise differences of −3, −1, and +3 percentage points. The full model exceeds Vision+Aux by 16.7 percentage points. Both variants use future fingertip changes as training targets. Vision+Aux receives fixed zero tactile input and estimates the current local state from visual history, whereas the full model directly observes the fingertip state. These observations provide a local reference for learning the relationship between robot actions and subsequent interaction changes. Together with the improvement over Vision+Touch, this comparison supports the complementary roles of current tactile feedback and auxiliary prediction in improving task success.

Taken together, these comparisons suggest that the auxiliary task helps the policy use tactile information by linking the observed local state to its subsequent changes under action. Joint optimization of the action-denoising and tactile-prediction losses encourages the shared temporal representation to capture information useful for both action generation and local dynamics prediction. This supervision shapes the tactile encoder and GRU during training. At inference, the policy uses the learned representation to generate actions from observation histories, while the auxiliary head is removed. The prediction task therefore contributes through representation learning without adding a separate prediction step during execution. The full model achieves the highest success rate across all three training seeds, supporting the effectiveness of combining tactile observations with auxiliary prediction.

## V. Conclusion

This paper presented a visuotactile diffusion policy with auxiliary prediction of rendered fingertip-image changes for simulated robotic assembly. A GRU aggregates visual and fingertip observation histories to condition diffusion-based action generation, while the training-only auxiliary head supplies action-conditioned supervision. Across three independent training seeds, each evaluated on 100 matched episodes, the full model achieves 59.0% ± 2.6% success, compared with 42.7% ± 1.5% for Vision-only, a gain of 16.3 percentage points. The exploratory seed-level 95% Student-t confidence interval for this gain is [12.5, 20.1] percentage points. Vision+Touch achieves 33.7% ± 0.6%, indicating that adding fingertip proxy input alone is insufficient in this configuration. The results support the combined design in the tested simulation setting. Future work will investigate auxiliary tactile-change prediction using real tactile sensors and evaluate the policy on a physical robot to examine sensor mismatch and transfer. It will also expand the number of independent training seeds and analyze representation quality and stage-specific failures.

## Acknowledgement

This work was kindly supported by the National Key R&D Program of China through grant 2025YFB4712702 and National Natural Science Foundation of China through grant No. 62403162.

## References

[1] G. Rabideau et al., “Planning, scheduling, and execution on the Moon: the CADRE technology demonstration mission,” in Proc. 24th Int. Conf. Autonomous Agents and Multiagent Systems (AAMAS), 2025.

[2] L. Burkhard et al., “Collaborative multi-rover crater exploration: Concept and results from the ARCHES analog mission,” in Proc. IEEE Aerospace Conf., 2024, doi: 10.1109/AERO58975.2024.10521301.

[3] P. Arm et al., “Scientific exploration of challenging planetary analog environments with a team of legged robots,” Science Robotics, vol. 8, no. 80, Art. no. eade9548, 2023, doi: 10.1126/scirobotics.ade9548.

[4] S. Tan, D. Zhou, X. Shao, J. Wang, and G. Sun, “Language-Conditioned Open-Vocabulary Mobile Manipulation with Pretrained Models,” in Proc. 34th Int. Joint Conf. Artificial Intelligence (IJCAI), pp. 8778–8786, 2025, doi: 10.24963/ijcai.2025/976.

[5] M. Lambeta et al., “DIGIT: A novel design for a low-cost compact high-resolution tactile sensor with application to in-hand manipulation,” IEEE Robotics and Automation Letters, vol. 5, no. 3, pp. 3838–3845, 2020, doi: 10.1109/LRA.2020.2977257.

[6] S. Wang, M. Lambeta, P.-W. Chou, and R. Calandra, “TACTO: A fast, flexible, and open-source simulator for high-resolution vision-based tactile sensors,” IEEE Robotics and Automation Letters, vol. 7, no. 2, pp. 3930–3937, 2022, doi: 10.1109/LRA.2022.3146945.

[7] C. Chi et al., “Diffusion Policy: Visuomotor policy learning via action diffusion,” in Proc. Robotics: Science and Systems (RSS), 2023, doi: 10.15607/RSS.2023.XIX.026.

[8] B. Huang, Y. Wang, X. Yang, Y. Luo, and Y. Li, “3D-ViTac: Learning fine-grained manipulation with visuo-tactile sensing,” in Proc. 8th Conf. Robot Learning, Proc. Machine Learning Research, vol. 270, pp. 2557–2578, 2025.

[9] H. Xue et al., “Reactive Diffusion Policy: Slow-fast visual-tactile policy learning for contact-rich manipulation,” in Proc. Robotics: Science and Systems (RSS), 2025, doi: 10.15607/RSS.2025.XXI.052.

[10] Open X-Embodiment Collaboration, “Open X-Embodiment: Robotic learning datasets and RT-X models,” in Proc. IEEE Int. Conf. Robotics and Automation (ICRA), pp. 6892–6903, 2024, doi: 10.1109/ICRA57147.2024.10611477.

[11] J. Ho, A. Jain, and P. Abbeel, “Denoising diffusion probabilistic models,” in Advances in Neural Information Processing Systems, vol. 33, 2020.

[12] J. Song, C. Meng, and S. Ermon, “Denoising diffusion implicit models,” in Proc. Int. Conf. Learning Representations (ICLR), 2021.

[13] A. Z. Ren et al., “Diffusion Policy Policy Optimization,” in Proc. Int. Conf. Learning Representations (ICLR), 2025.

[14] R. Cadene et al., “LeRobot: State-of-the-art machine learning for real-world robotics in PyTorch,” software, 2024. Implementation revision e595b7902714ba51f91e47523f66f89c5181b649.

[15] A. Mandlekar, D. Xu, R. Martín-Martín, Y. Zhu, L. Fei-Fei, and S. Savarese, “Human-in-the-loop imitation learning using remote teleoperation,” arXiv:2012.06733, 2020.

[16] Z. He, H. Fang, J. Chen, H.-S. Fang, and C. Lu, “FoAR: Force-Aware Reactive Policy for Contact-Rich Robotic Manipulation,” arXiv:2411.15753, 2024.

[17] M. Wang, Z. Yue, R. Lu, Y. Li, Z. Wang, G. Pan, K. Dong, J. Cheng, Y. Cheng, and H. Liu, “PhaForce: Phase-Scheduled Visual-Force Policy Learning with Slow Planning and Fast Correction for Contact-Rich Manipulation,” arXiv:2603.08342, 2026.

[18] J. Li, T. Wu, J. Zhang, Z. Chen, H. Jin, M. Wu, Y. Shen, Y. Yang, and H. Dong, “Adaptive Visuo-Tactile Fusion with Predictive Force Attention for Dexterous Manipulation,” arXiv:2505.13982, 2025.

[19] K. He, X. Zhang, S. Ren, and J. Sun, “Deep Residual Learning for Image Recognition,” in Proc. IEEE Conf. Computer Vision and Pattern Recognition (CVPR), pp. 770–778, 2016, doi: 10.1109/CVPR.2016.90.

[20] C. Finn, X. Y. Tan, Y. Duan, T. Darrell, S. Levine, and P. Abbeel, “Deep Spatial Autoencoders for Visuomotor Learning,” in Proc. IEEE Int. Conf. Robotics and Automation (ICRA), pp. 512–519, 2016, doi: 10.1109/ICRA.2016.7487173.

[21] E. Perez, F. Strub, H. de Vries, V. Dumoulin, and A. C. Courville, “FiLM: Visual Reasoning with a General Conditioning Layer,” in Proc. AAAI Conf. Artificial Intelligence, vol. 32, no. 1, pp. 3942–3951, 2018, doi: 10.1609/AAAI.V32I1.11671.